\documentclass[10pt]{article}

\usepackage{times}
\usepackage{amsmath}
\usepackage{amssymb}
\usepackage{cite}
\usepackage{bm}
\usepackage{float}
\usepackage{url}
\usepackage[ruled]{algorithm}
\usepackage{algorithmicx}
\usepackage{algpseudocode}
\usepackage{graphicx}
\usepackage{subfigure}
\usepackage[small]{caption}
\usepackage{color}
\usepackage{graphics}
\usepackage{graphicx,epsfig,psfrag,pstricks}
\usepackage{multirow}
\usepackage{array}

\title{Generalized Wishart processes for interpolation 
	over diffusion tensor fields}
\author{Hern\'an Dar\'io Vargas Cardona, Mauricio A. \'Alvarez and Alvaro A. Orozco\\
{\small \emph{Faculty of Engineering, Universidad Tecnol{\'o}gica de Pereira, Colombia, 660003.}}\\}
\date{}

\begin{document}
\maketitle

\begin{abstract}
Diffusion Magnetic Resonance Imaging (dMRI) is a non-invasive tool for
watching the microstructure of fibrous nerve and muscle
tissue. From dMRI, it is possible to estimate 2-rank diffusion tensors
imaging (DTI) fields, that are widely used in clinical applications:
tissue segmentation, fiber tractography, brain atlas construction,
brain conductivity models, among others. Due to hardware limitations
of MRI scanners, DTI has the difficult compromise between spatial
resolution and signal noise ratio (SNR) during acquisition. For this
reason, the data are often acquired with very low resolution. To
enhance DTI data resolution, interpolation provides an interesting
software solution. The aim of this work is to develop a methodology
for DTI interpolation that enhance the spatial resolution of DTI
fields. We assume that a DTI field follows a recently introduced
stochastic process known as a generalized Wishart process (GWP), which
we use as a prior over the diffusion tensor field. For posterior inference, we use
Markov Chain Monte Carlo methods. We perform experiments in toy and real data. Results
of GWP outperform other methods in the literature, when 
compared in different validation protocols.
\end{abstract}

\section{Introduction}

Diffusion Magnetic Resonance Imaging (dMRI) is a non-invasive
procedure to find connections into biological mediums such as fiber nerves
and muscle tissue. From dMRI it is possible to
estimate the apparent diffusivity coefficient (ADC) of water particles
within tissue by solving the Stejskal-Tanner formulation
\cite{Basser1994}. 
A 2-rank diffusion tensor $(D)$ is employed to
modeling ADC in each specific voxel, where $D$ is a symmetric and positive
definite $3\times3$ matrix. Following this notion, a
diffusion tensor imaging (DTI) field is understood as a grid of
individual but related diffusion tensors.  Although a DTI field shows
how some nerve fiber bundles are interconnected, there are some
limitations in MRI scanners. For example, dMRI is sensitive to the
difficult compromise between spatial resolution and signal to noise ratio
(SNR). This leads to data acquisitions with low resolution
\cite{Yang2014}.

To enhance DTI data resolution, interpolation provides an interesting
and feasible methodological solution. Diffusion tensor fields belong
to a Riemannian space, where the Riemannian metric is defined by the
inner product assigned to each point of this space. With this metric,
one can to compute geodesic distances between diffusion tensors and to
calculate different statistics in this space \cite{Fletcher2007}. An
important condition is keeping the smooth transition of anisotropic
features inherent in the given tensor fields (i.e. Fractional
anisotropy-FA maps), especially around degenerate
points, where at least two of three eigenvalues are equivalent
\cite{Bi2010}. Interpolation of the diffusion tensor fields have many
applications. For example, registration of DTI datasets will require
resolution enhancement when a registration transformation is applied
to a tensor field. Other examples that require DTI interpolation
include segmentation, atlas construction, diagnosis of neurological
diseases, etc \cite{Barmpoutis2007}. Currently, the clinical
acquisition protocols of dMRI data allow one or two millimeters
resolution for each voxel. The problem here, is that brain tissue
fiber bundles are in micrometers scale. Therefore, the tractography
models developed from DTI data can be imprecise due to the current low
resolution in acquired images. Normally, visualization of DTI is
discrete, where it is used ellipsoids or glyphs for graphic
representation. Tractography is the search of fiber
connection among neighboring voxels. The basic idea
is to generate a continuous data representation
\cite{Hotz2010}. According to this, an accurate interpolation approach
may improve the spatial resolution of diffusion tensor fields in a
considerable factor. Therefore, the tractography process
will describe with more detail the fiber tissue connection.

Some recent works have proposed interpolation methods for tensor
fields in DTI. They developed a variety of mathematical approaches,
such as: direct smooth approximation \cite{Pajevic2002} and euclidean
approaches, but they do not retain the principal properties of a DTI,
i.e positive definite tensors. For this reason, the scientific
community has been looking alternative methods for estimating tensor
fields that keep the symmetric positive definite (SPD) constraint
inside the grid of tensors. \cite{Arsigny2006} presented a Log-euclidean approximation
and \cite{Fletcher2007} developed a Riemannian framework
achieving important advances in tensor fields geometry, but they lack
in smoothness property in presence or high level of
noise. \cite{Barmpoutis2007} presented a b-spline scheme that
interpolates SPD tensors with high accuracy using the Riemannian
metric. The authors introduced a tensor product of B-splines that
minimizes the Riemannian distance between tensors. Following the
Riemannian framework, \cite{Kindlmann2007} presented
Geodesic-loxodromes that can identify isotropic and anisotropic
components of the tensor and interpolates each component
separately. Finally, alternative methodologies have been posited: a
tensor field reconstruction based on eigenvector and eigenvalue
interpolation \cite{Hotz2010}, location of degenerated lines in 2-D
planar \cite{Bi2010}, and a feature-based interpolation
\cite{Yang2012}. However, those methods do not achieve an adequate
representation of a DTI field obtained from noisy real data.

As previously was pointed out, most of the methods for DTI interpolation are
based on Riemannian geometry. While they preserve the main properties
of DTI data, and solve limitations of the Euclidean approaches, they 
lead to rigid interpolations that fail to fully adapt to the
variety of diffusion patterns in biological tissues
\cite{Chang2012}. In this work, we present a novel methodology for
interpolation of DTI fields. Instead of a Riemann geometry framework,
we propose a stochastic modeling of DTI. We assume that a DTI field
follows a generalized Wishart process (GWP). A GWP is a collection of
symmetric positive definite random matrices indexed by an arbitrary
dependent variable \cite{Wilson2011}, i.e. the $x,y,z$ position. In
this context, we use it to model the entire DTI field
$D(x,y,z)$. Then, through approximate Bayesian inference (i.e
Elliptical slice sampling and Markov Chain Monte Carlo methods), we
estimate the optimal parameters of the model. Stochastic modeling of DTI fields has some
advantages: positive definite matrices, robustness to noise, smooth transition among nearby tensors and
good accuracy for estimating new data. We compare our approach
with linear interpolation \cite{Pajevic2002} and a Riemannian method known as log-euclidean interpolation 
\cite{Arsigny2006}. We perform experiments in toy and real DTI
data. Results of GWP improve to the comparison methods in different
validation protocols.

\section{Materials and methods}
\subsection{DTI estimation from dMRI and DTI fields}
Diffusion Magnetic Resonance Imaging (dMRI)
studies the diffusion of water particles in the human brain. 
Diffusion can be described by a symmetric positive definite $3\times3$
matrix proportional to the covariance of a Gaussian distribution \cite{Basser1994,ST1965}.

\small{
	\begin{equation*}
	\textbf{D}=\begin{bmatrix}{D_{xx}}&{D_{xy}}&{D_{xz}}\\{D_{yx}}&{D_{yy}}&{D_{yz}}\\{D_{zx}}&{D_{zy}}&{D_{zz}}\end{bmatrix}
	\label{ec4}
	\end{equation*}
}

\normalsize
For water, the diffusion tensor (DT) is symmetric, so that $ D_ {ij} =
D_ {ji} $, where $ i, j = x, y, z $.
The diffusion tensor for each voxel of the dMRI is calculated using
the Stejskal-Tanner formulation \cite{Basser1994}:
\small{
	\begin{equation}
	S_k=S_0e^{-b\hat{\bold{g}}_k^\top D\hat{\bold{g}}_k},
	\label{eq:ST}
	\end{equation}
}

\normalsize
where $S_k$ is the $k^{th}$ dMRI, $S_0$ is the reference image,
$\hat{\bold{g}}_k$ is the gradient vector and $b$ is the diffusion
coefficient. At least $7$ dMRI measurements are necessary for each
slice ($k=0,1,...,7$). Usually, DTI fields are estimated from \eqref{eq:ST} using least
squares \cite{Basser1995}. However, there are robust methods for DT
estimation. In this work, we use the RESTORE algorithm
\cite{Restore2005} for solving the DTs.

Traditionally, rank-2 DTs have been visualized by constructing the ellipsoid given by:

\small{
	\begin{equation}
	\mathbf{r}^\top D^{-1}\mathbf{r}=C
	\label{eq:ellpisoid}
	\end{equation}
}
\normalsize
where $\mathbf{r}^\top =
[x, y, z]$ is the position vector, and $C$ is a constant with
the units of time. Therefore, the resulting shape is a level
surface of the expression on the left side of \eqref{eq:ellpisoid}, and
it is possible to show by diagonalization that these surfaces
are ellipsoids. 


\subsection{Generalized Wishart Process (GWP)}

We begin with the Wishart distribution, which defines a probability
density function over a symmetric positive definite matrix. Let $S$ be
a $p\times p$ symmetric positive definite matrix of random
variables. Let $V$ be a (fixed) positive definite matrix of size
$p\times p$. Then, if $\nu \geq p$, $S$ has a Wishart distribution
with $\nu$ degrees of freedom if it has a probability density function
given by:

\small{
	\begin{equation*}
	S=\frac{|S|^{\nu-p-1}}{2^{\nu p/2}|V|^{\nu/2}\Gamma_p(\frac{\nu}{2})}e^{-\frac{1}{2}\operatorname{trace}(V^{-1}S)},
	\end{equation*}
}
\normalsize
where $|\cdot|$ is the determinant and $\Gamma_p$ is the multivariate gamma function:

\small{
	\begin{equation*}
	\Gamma_p\left( \frac{\nu}{2}\right)=\pi^{\frac{p(p-1)}{4}}\prod_{j=1}^{p}\Gamma\left(\frac{\nu}{2}+\frac{1-j}{2} \right) 
	\end{equation*}
}
\normalsize
Following this notion and according to the definition given in
\cite{Wilson2011}, a generalized Wishart process (GWP) is as a
collection of symmetric positive definite random matrices indexed by
an arbitrary and high dimensional dependent variable $\mathbf{z}$. In
DTI fields, the dimension is $p=3$ because diffusion tensors are
represented by $3\times3$ matrices, and the indexed variable refers to
position coordinates $\textbf{z}=[x,y,z]^{\top}$. Assume $3\nu$ independent
Gaussian process functions $u_{id}(\mathbf{z})\sim \mathcal{GP}(0,k)$, for 
$i=1,...,\nu$ and $d=1,2,3$, where $k(\mathbf{z},\mathbf{z}')$ is the covariance or kernel
function for the GP. Given a set of input vectors $\{\mathbf{z}\}_{n=1}^N$, the vector
$(u_{id}(\mathbf{z}_1),u_{id}(\mathbf{z}_2)
,...,u_{id}(\mathbf{z}_N))^\top \sim \mathcal{N}(\mathbf{0},K)$,
being $K$ an $N\times N$ Gram matrix with entries
$K_{ij}=k(\mathbf{z}_i,\mathbf{z}_j)$. If we define
$\hat{\mathbf{u}}_i(\mathbf{z})=\left(u_{i1}(\mathbf{z}),u_{i2}(\mathbf{z}),u_{i3}(\mathbf{z})\right)^\top$
and $L$ as the lower Cholesky decomposition of a $p\times p$ scale
matrix $V$, such that $LL^\top=V$, for each input position
$\mathbf{z}=[x,y,z]^{\top}$, the diffusion tensor $D(\mathbf{z})$ follows a Wishart
distribution,

\small{
	\begin{equation}
	D(\mathbf{z})=\sum_{i=1}^{\nu} L \hat{\mathbf{u}}_i(\mathbf{z}) \hat{\mathbf{u}}^\top_i(\mathbf{z}) L^\top \sim \mathcal{GWP}_p(\nu,V,k(\cdot,\cdot)),
	\label{GWP}
	\end{equation}
}
\normalsize
In this work, we use the squared exponential kernel $k(\mathbf{z},\mathbf{z'})$,

\small{
	\begin{equation*}
	k(\mathbf{z},\mathbf{z'})=\exp \left(-0.5\frac{ \lVert \mathbf{z}-\mathbf{z'} \rVert^2}{\theta^2} \right),
	\end{equation*}
}
\normalsize
where $\theta$ is the length-scale hyperparameter.

\subsection{Bayesian inference for DTI field learning}

In order to perform DTI interpolation, we first need to compute the posterior distribution for the 
variables in the model. 
For a DTI field, we assume a prior given by a Generalized Wishart process
\small{
	\begin{equation}
	p\left( D(\mathbf{z})\right)  \sim \mathcal{GWP}_3(\nu,V,k(\cdot,\cdot))=
	\displaystyle \sum_{i=1}^\nu L \hat{\mathbf{u}}_i(\mathbf{z})\hat{\mathbf{u}}_i^\top(\mathbf{z})L^\top.
	\label{prior}
	\end{equation}
}
\normalsize
For the likelihood function, we assume each element from the diffusion tensor data 
follows an independent Gaussian distribution with the same variance $\sigma^2$. This leads to a likelihood with the following form:
\small{
	\begin{equation*}
	p(S|u,L,\nu) \propto \prod_{i=1}^{N}\exp\left(-\frac{1}{2\sigma^2} \lVert S(\mathbf{z}_i)- D(\mathbf{z}_i)
	\rVert_{frob}^2 \right), 
	\end{equation*}
}
\normalsize
where $S(\mathbf{z})$ is the known initial DTI field with low resolution, $D(\mathbf{z})$ is constructed from equation \eqref{prior}, and Frobenius norm is given by
\small{
	\begin{equation*}
	\lVert \mathbf{X} \rVert_{frob}^2=\operatorname{trace}\left(\mathbf{X}^T\mathbf{X} \right). 
	\end{equation*}
}
\normalsize
The purpose is to infer the posterior probability of $D(\mathbf{z})$
given a known tensorial data set
$S(\mathbf{z})=\{S(\mathbf{z}_1),S(\mathbf{z}_2),...,S(\mathbf{z}_N)\}$,
being $N$ the number of data in the initial DTI field. We first
compute the posterior of the relevant variables in equation
\eqref{prior} including the vector of all GP function values $\mathbf{u}$,
length-scale hyperparameter of the GP kernel function $\theta$, the
lower Cholesky decomposition of the scale matrix $L$, such that $LL^\top=V$, and the
degrees of freedom $\nu$. Given a GWP prior for the model and the likelihood function, the posterior distributions
can be computed by 

\small{
	\begin{align}
	p(\mathbf{u}|\theta,L,S)\propto p(S|\mathbf{u},L,\nu)p(\mathbf{u}|\theta), \label{eq:u}\\
	p(\theta|\mathbf{u},L,S)\propto p(\mathbf{u}|\theta,L,D)p(\theta), \label{eq:theta}\\
	p(L|\mathbf{u},\theta,S)\propto p(S|\mathbf{u},L,\nu)p(L). \label{eq:L}
	\end{align}
}
\normalsize
We use Markov chain Monte Carlo algorithms to sample in cycles. We
employ Metropolis-Hastings to sample $\theta$ from \eqref{eq:theta}, and the elements of
scale matrix $L$ from \eqref{eq:L}. To sample $\mathbf{u}$ from \eqref{eq:u}, we employ elliptical slice
sampling \cite{Murray2010}. We choose $\nu=5$ through
cross-validation. We set a log-normal prior on $\theta$, a spherical
Gaussian prior on elements of $L$ and the prior
$p(\mathbf{u}|\theta)\sim \mathcal{N}(\mathbf{0},K_B)$ is a Gaussian
distribution with $3\nu N\times3\nu N$ block diagonal covariance
matrix $K_B$, formed using $3\nu$ of the $K$ matrices.

\subsection{DTI field interpolation through GWP modeling}

Once we find the posterior distributions over all relevant variables for the model,
we can compute the posterior distribution for
$D(\mathbf{z}_*)$ in a new spatial position
$\mathbf{z}_*=[x_*,y_*,z_*]^{\top}$. First, we have to infer the distribution
over all unknown GP function values $\mathbf{u}_*$ in $\mathbf{z}_*$, where 
$\mathbf{u}_*$ is a vector with elements given by $u_{id}(\mathbf{z}_*)$.
The joint distribution over $\mathbf{u}$ and $\mathbf{u}_*$ is given by,
\small{
	\begin{equation*}
	\begin{bmatrix}{\mathbf{u}}\\{\mathbf{u}_*}\end{bmatrix}\sim \mathcal{N}\left(\mathbf{0},\begin{bmatrix}{K_B}&{A^\top}\\{A}&{I_p}\end{bmatrix} \right) 
	\end{equation*}
}
\normalsize
If $\mathbf{u}_*$ and $\mathbf{u}$ have $p$ and $q$ elements respectively, 
$A$ is a $p\times q$ matrix that represents the covariances between $\mathbf{u}_*$
and $\mathbf{u}$ for all pairs of training and validation data, this is $A_{ij}=k_i(\mathbf{z}_*,
\mathbf{z}_j)$ for $i+(i-1)N\leq j\leq iN$, and $0$ otherwise. $I_p$ is a $p\times p$ identity matrix. Using the properties of a Gaussian distribution, and conditioning on $\mathbf{u}$, we obtain:

\small{
	\begin{equation}
	p\left( \mathbf{u}_*|\mathbf{u}\right) \sim \mathcal{N}\left(AK_B^{-1}\mathbf{u},I_p-AK_B^{-1}A^\top \right)
	\label{post} 
	\end{equation}
}
\normalsize
From values of $\mathbf{u}_*$ obtained from \eqref{post}, 
and using equation \eqref{GWP}, we can construct $D(\mathbf{z}_*)$.

\subsection{Validation procedure and datasets}
As ground truth (gold standard) we employ three different types of
data. The first one corresponds to a synthetic DTI field. The second
one corresponds to a simulation of crossing fibers using the
algorithm of the fanDTasia toolBox \cite{Barmpoutis2010}, available at
\url{http://www.cise.ufl.edu/~abarmpou/lab/fanDTasia/}. The third one,
corresponds to a DTI dataset estimated from real dMRI through the
RESTORE method \cite{Restore2005}. dMRI data of the head were acquired
from a healthy subject on a General Electric Signa HDxt 3.0T MR
scanner using the body coil for excitation, and an 8-channel
quadrature brain coil for reception.  We employ $25$
gradient directions with a value for b equal to $1000$ $S/mm^2$. The
study contains $128\times128\times33$ images in axial plane. For the
three datasets, we downsample the DTI field by a factor of two. The
downsampled field is the input data for the GWP. After we perform inference over the
GWP, we interpolate the DTI field, and calculate two error
metrics, having the gold standards as our references. Also, we repeat
the same procedure for linear \cite{Pajevic2002} and log-euclidean interpolation \cite{Arsigny2006} for a
comparison with two commonly used methods in the state of the
art. We use two metrics to measure the differences between the
interpolated fields and the ground truth, the Frobenius norm, and the Riemman
distance, defined by
\small
\begin{align*}
\operatorname{Frob}(T_1,T_2)=\sqrt{\operatorname{trace}\left[ \left(T_1-T_2\right)^\top \left(T_1-T_2\right)  \right] },  \\
\operatorname{Riem}(T_1,T_2)=\sqrt{\operatorname{trace}\left[  \log(T_1^{-1/2} T_2 T_1^{-1/2})^\top \log(T_1^{-1/2} T_2 T_1^{-1/2}) \right] }, 
\end{align*}
\normalsize
where $T_1$ and $T_2$ are the estimated and the ground truth tensors, respectively.
The error metrics are computed for each \textit{voxel}. 
We report the mean and standard deviation for the errors over the predicted data.

\section{Experimental results and discussion}
In this section, we present the interpolation results for the different DTI datasets. We compare 
with linear \cite{Pajevic2002} and log-euclidean interpolation \cite{Arsigny2006}.

\subsection{Synthetic Data}
\begin{figure}[H]
	\centering
	\subfigure[\label{st_t} ]{\includegraphics[scale=0.138]{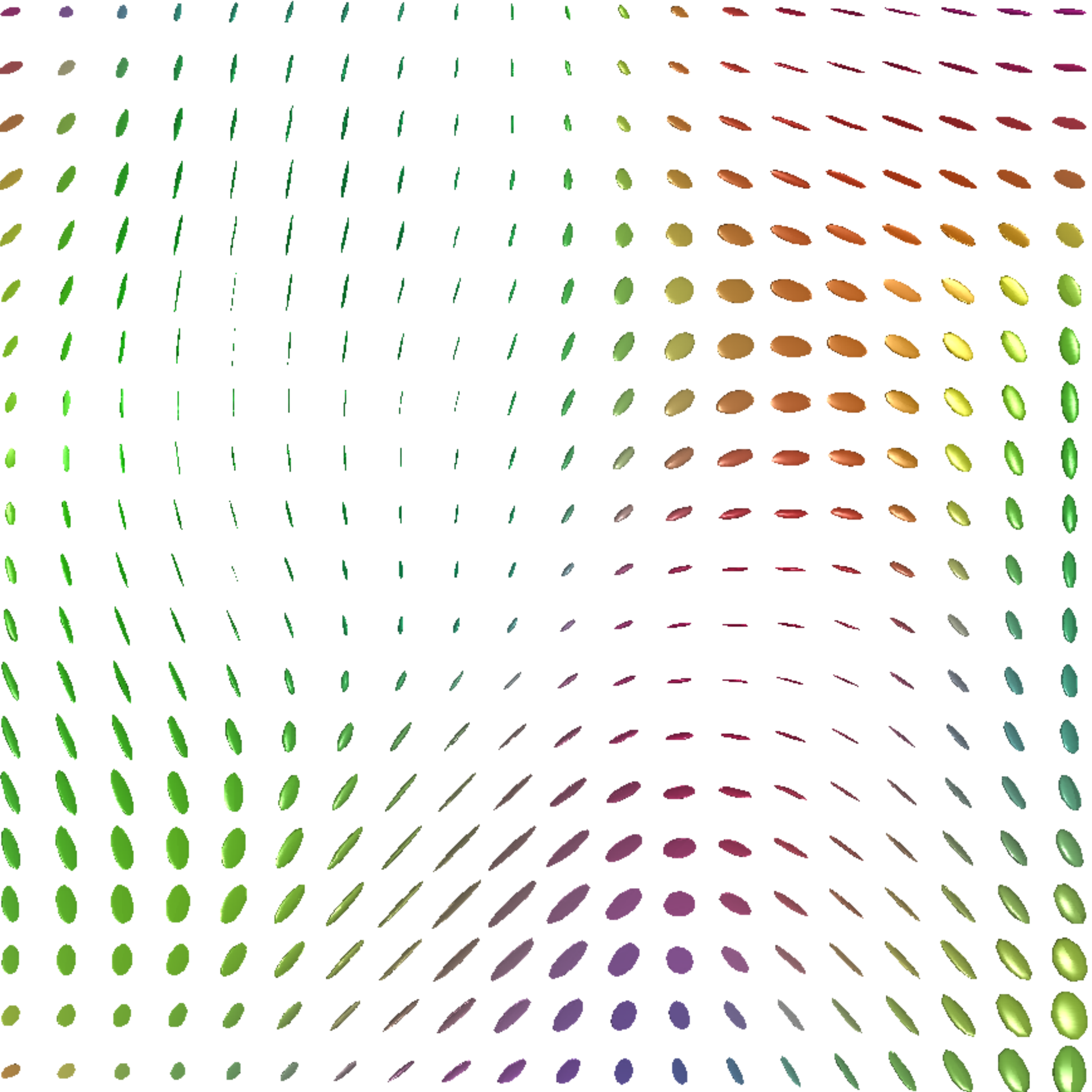}}
	\subfigure[\label{st_lin}  ]{\includegraphics[scale=0.23]{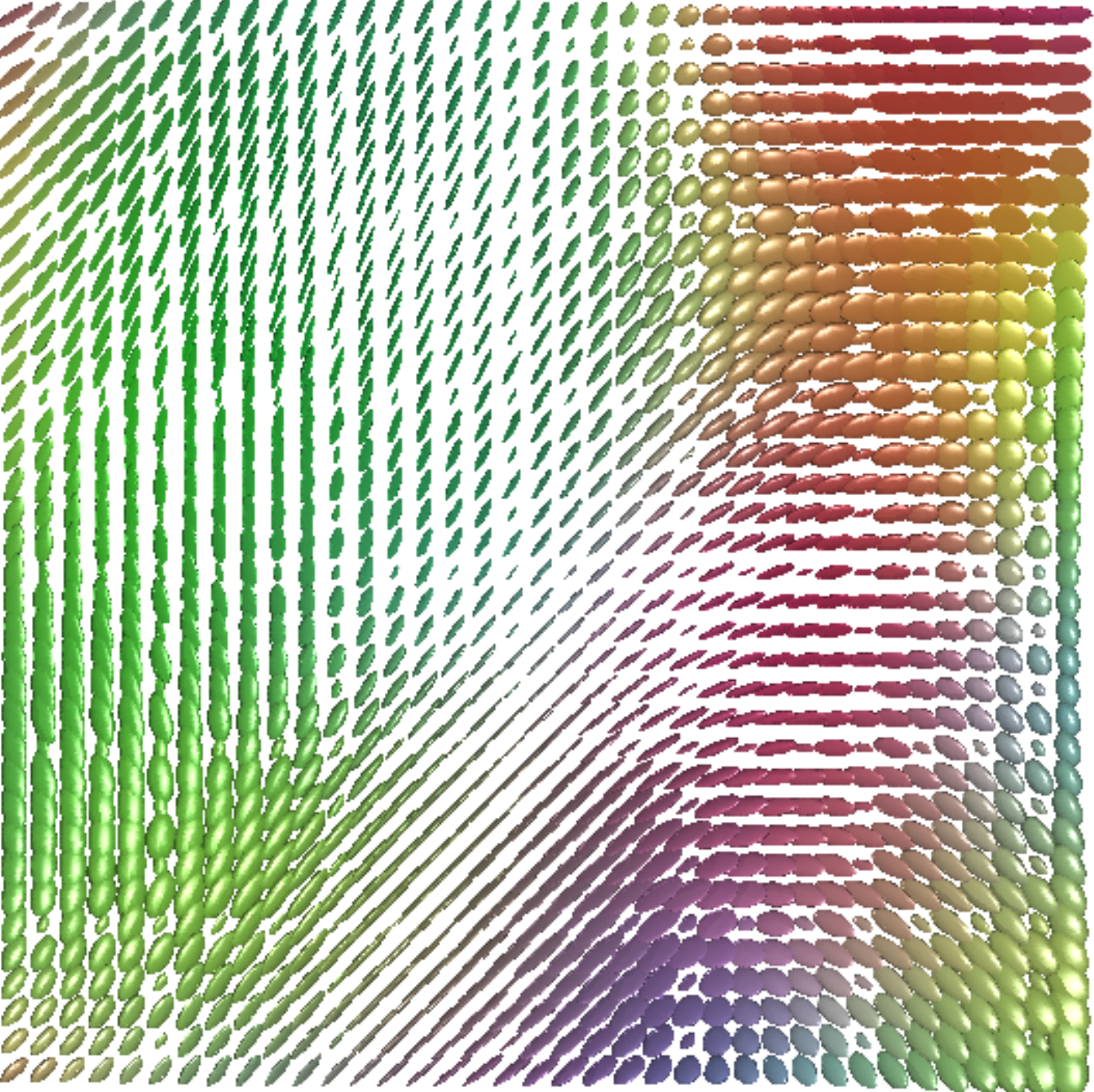}}
	\subfigure[\label{st_log} ]{\includegraphics[scale=0.23]{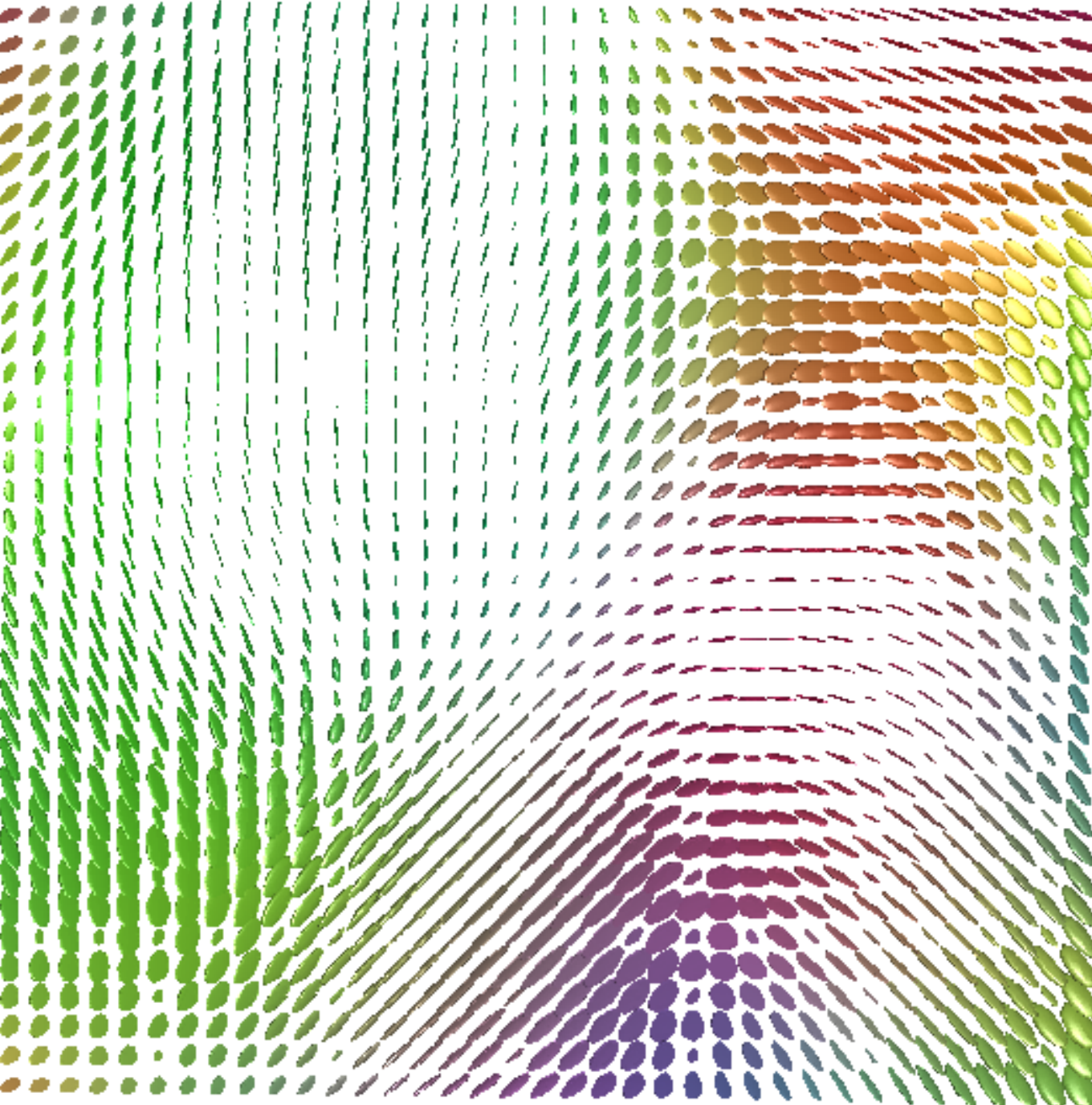}}	
	\subfigure[\label{st_gwp}]{\includegraphics[scale=0.138]{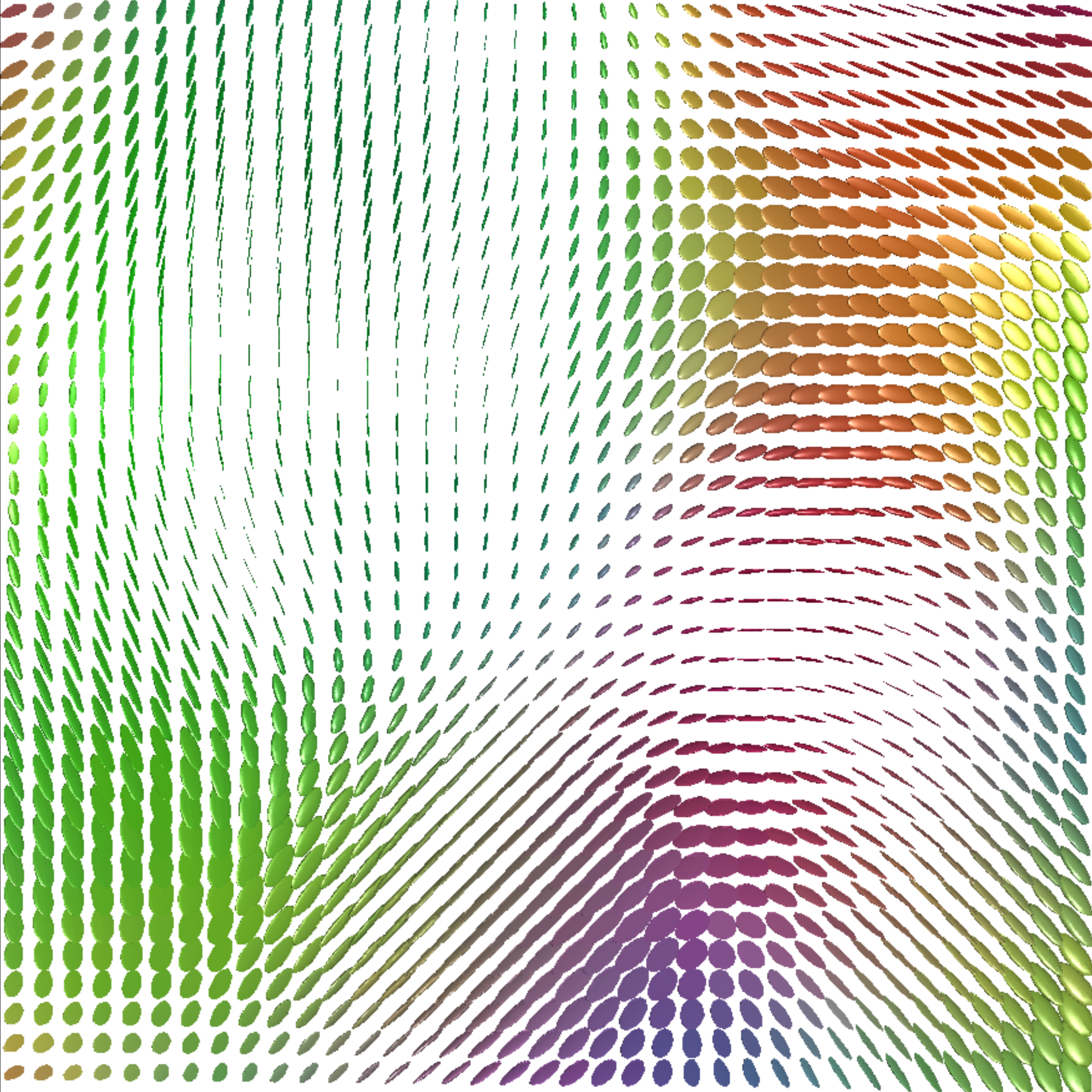}}
	\subfigure[\label{st_gt} ]{\includegraphics[scale=0.138]{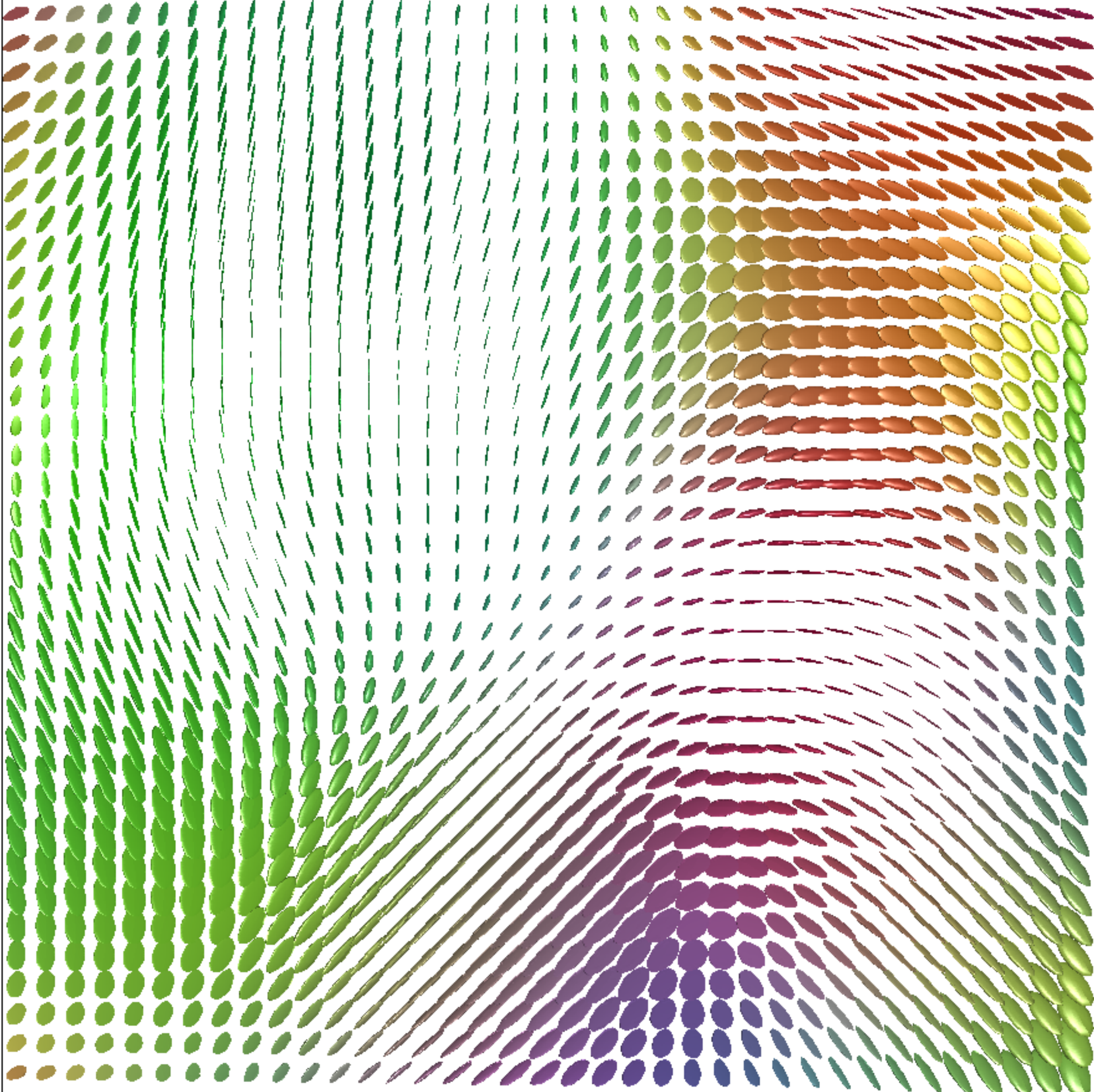}}
	\caption{Graphic results for DTI interpolation $(2\times)$ applied in synthetic data. (a) Downsampled DTI field (data used for estimation). (b) Linear interpolation. (c) Log-euclidean interpolation. (d) Interpolation with GWP. (e) Ground truth. }	
	\label{f_st}
\end{figure}

We generate noisy random DTI data to construct a $2D$ field of
$37\times 37$ tensors. We assume $25$ gradient directions for
generating DTs, and $b$ value of $1000$ $s/mm^2$. In Figure \ref{f_st} we
can see the initial downsampled DTI field, linear and log-euclidean
interpolation, the interpolated field with GWP, and the ground truth
respectively. Table \ref{tabR1} shows the error metrics.

\begin{table}[H]
	\centering
	\caption{Metric results for synthetic DTI field}
	\begin{tabular}{ccc}
		\hline
		& Frobenius distance $(\times10^{-5})$ & Riemman distance \\
		\hline
		GWP   & $7.06\pm1.51 $ & $0.160\pm0.125 $ \\
		linear interpolation &   $50.11\pm4.26$    & $8.54\pm1.36$  \\
		log-euclidean &  $35.25\pm3.92$ & $6.34\pm1.22$ \\
		\hline
	\end{tabular}%
	\label{tabR1}%
\end{table}%

\subsection{DTI from crossing fibers}
\begin{figure}[H]
	\centering
	\subfigure[\label{cf_t} ]{\includegraphics[scale=0.13]{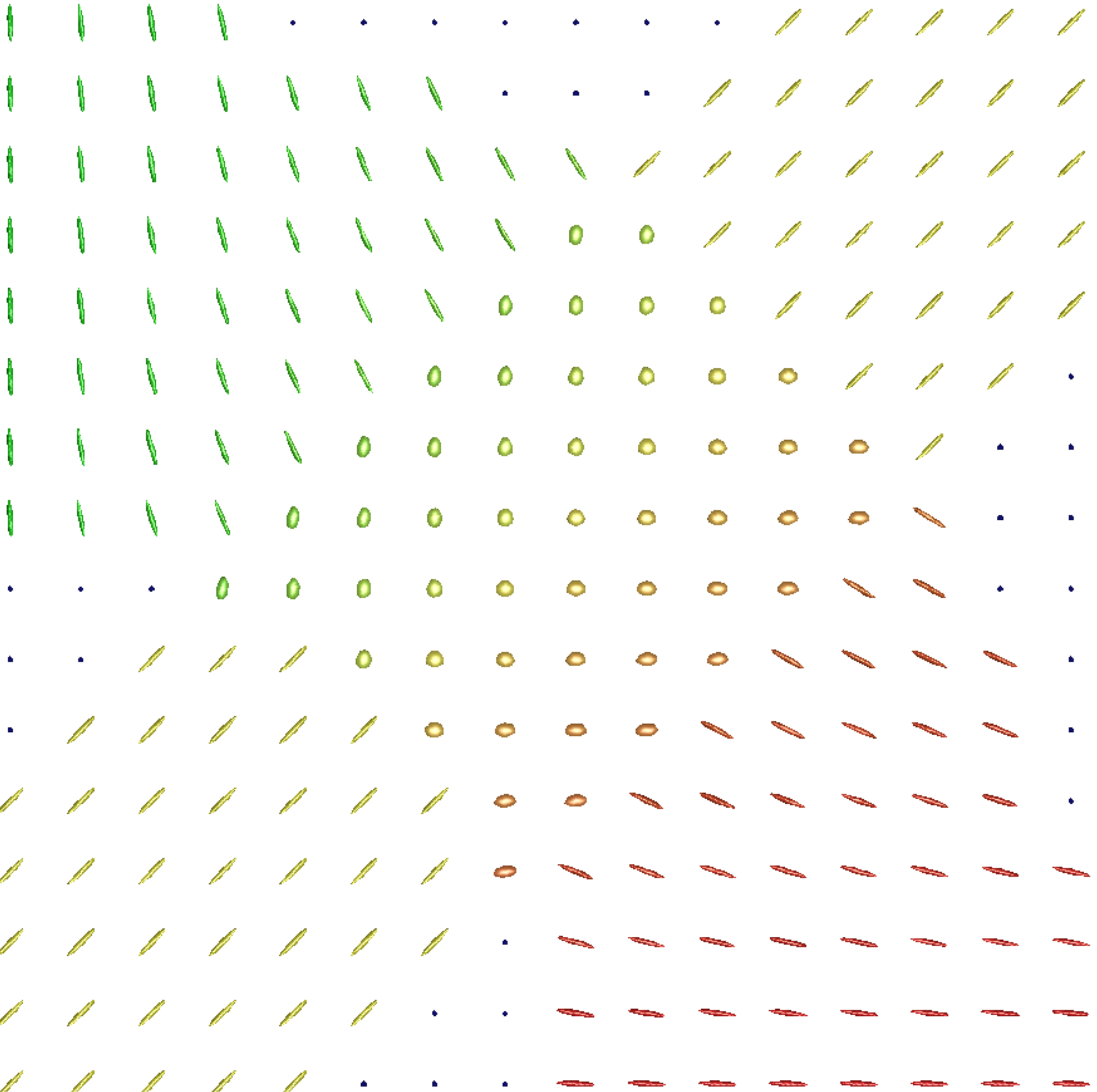}}
	\subfigure[\label{cf_lin} ]{\includegraphics[scale=0.215]{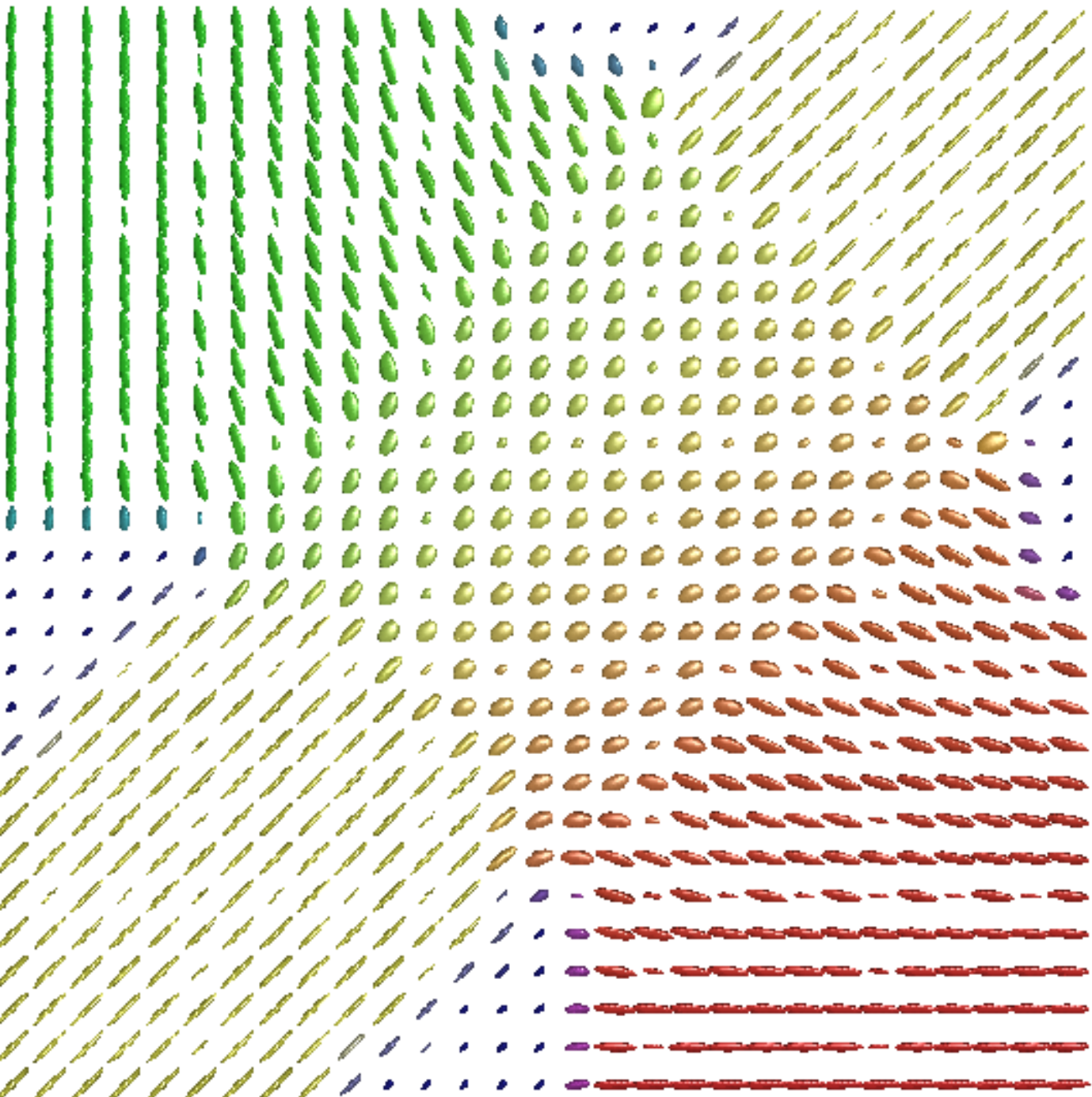}}
	\subfigure[\label{cf_log} ]{\includegraphics[scale=0.228]{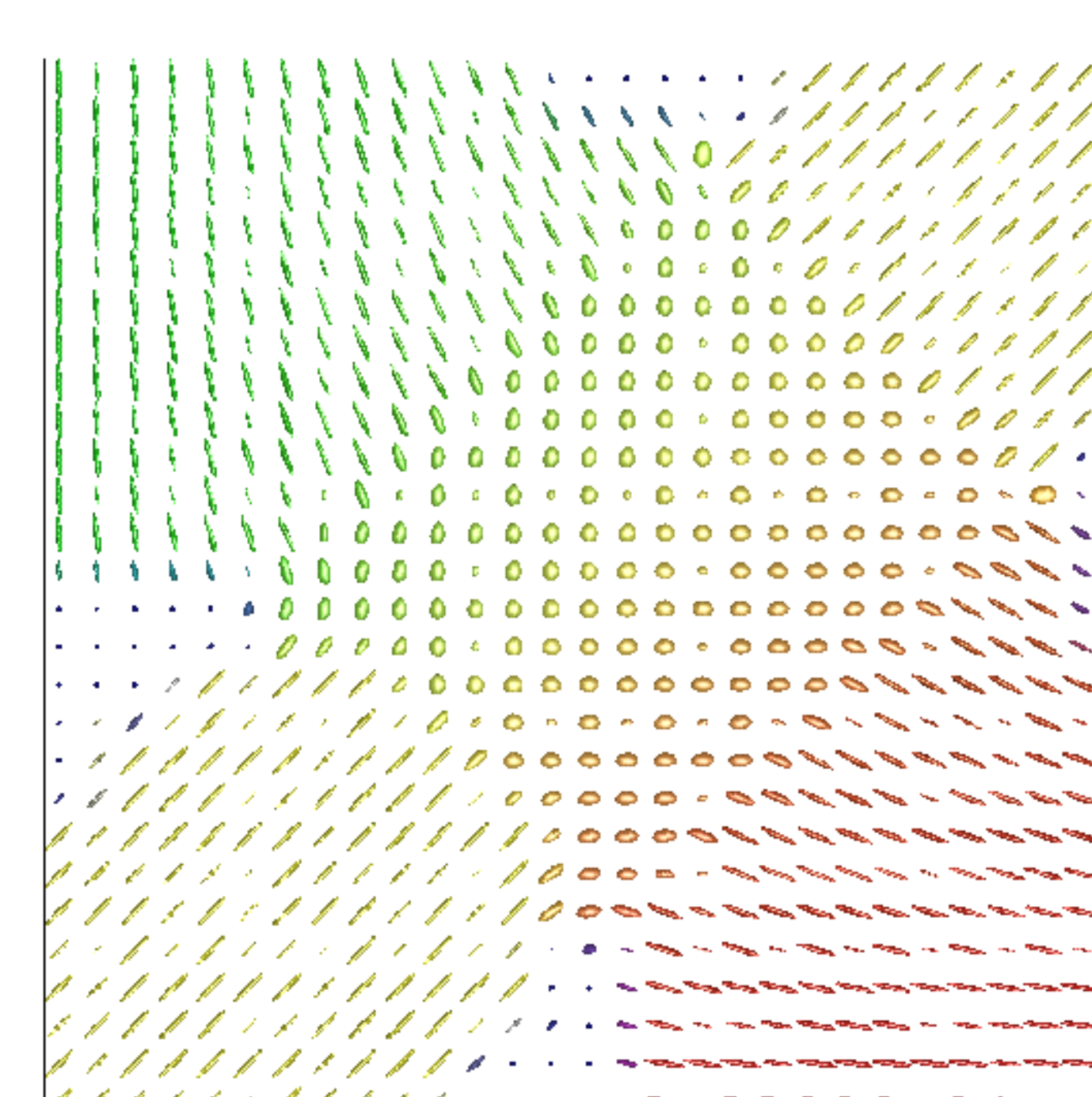}}
	\subfigure[\label{cf_gwp}]{\includegraphics[scale=0.13]{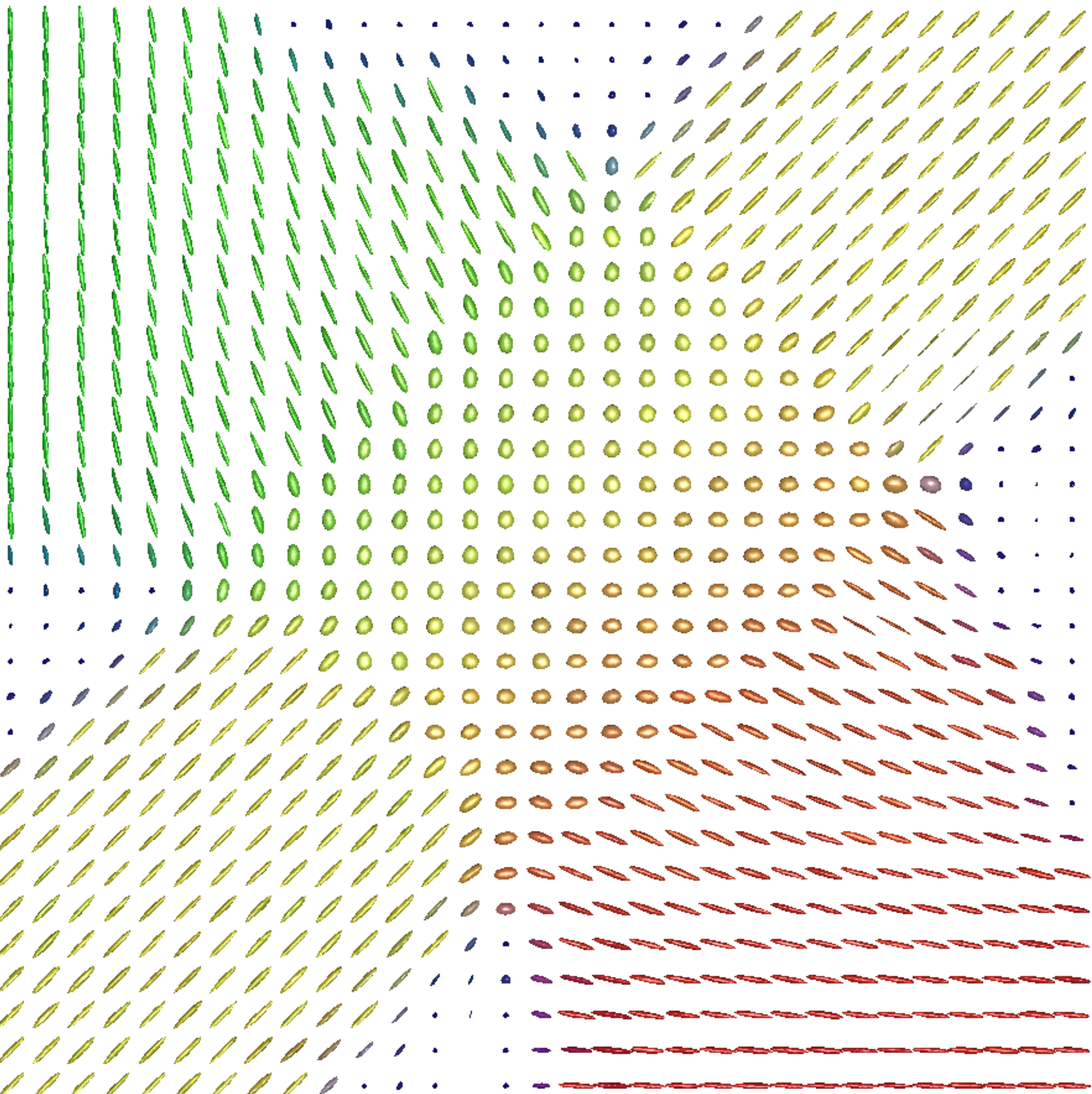}}
	\subfigure[\label{cf_gt} ]{\includegraphics[scale=0.13]{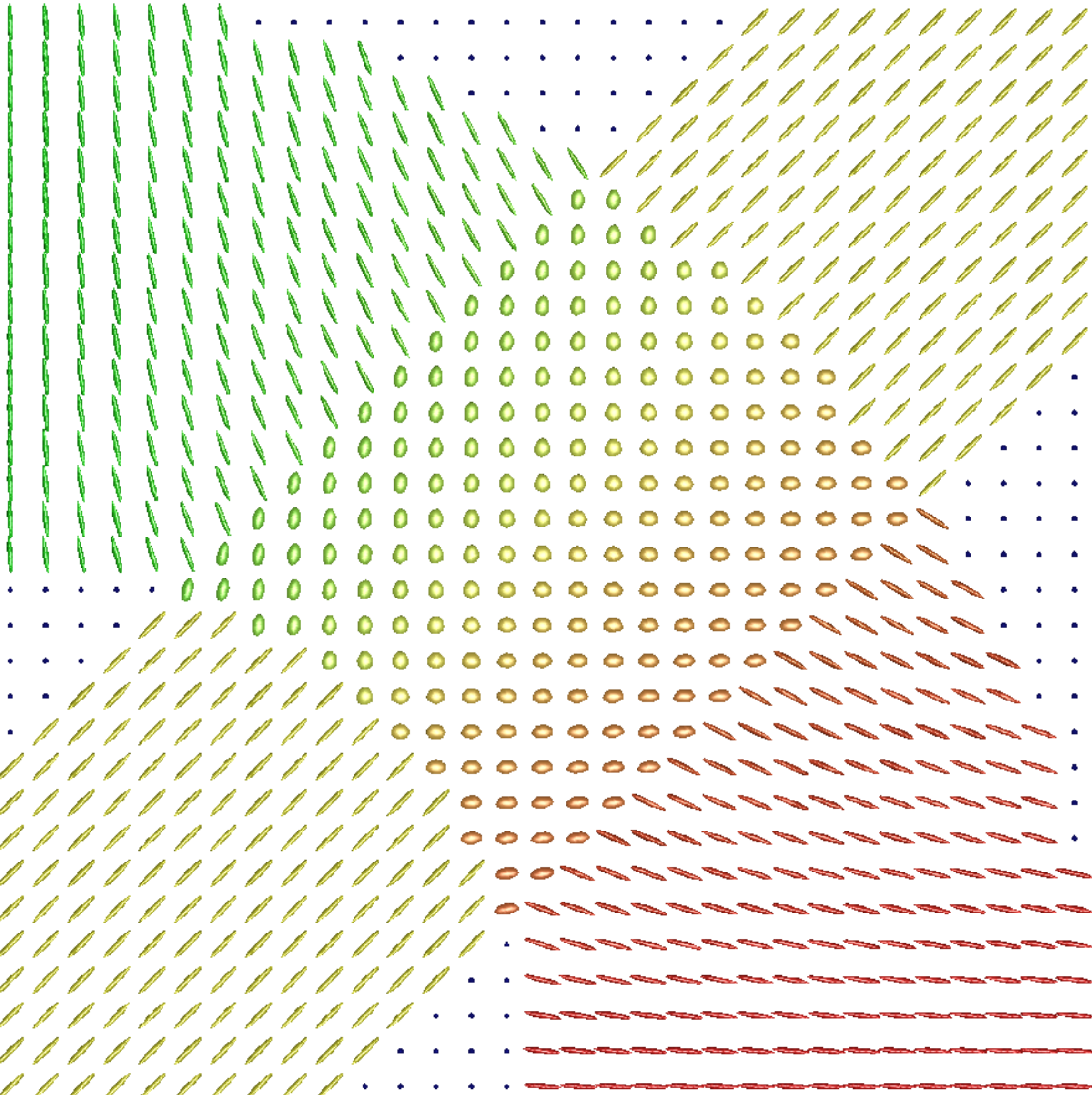}}
	\caption{Graphic results for DTI interpolation $(2\times)$ applied in crossing fibers field. 
		(a) Downsampled DTI field (data used for estimation). (b) Linear interpolation. (c) Log-euclidean interpolation. (d) Interpolation with GWP. (e) Ground truth.}
	\label{f_cb}	
\end{figure}

One of the most critical DTI datasets correspond to crossing
fibers. We generate this type of DTI field through FanDTasia toolbox
\cite{Barmpoutis2010}. This dataset describes a $2D$ crossing fiber
field with $31\times31$ tensors. Figure \ref{f_cb} and Table
\ref{tabR2} show the comparative results.

\begin{table}[H]
	\centering
	\caption{Error measures for crossing fibers in a DTI field.}
	\begin{tabular}{ccc}
		\hline
		& Frobenius distance $(\times10^{-5})$ & Riemman distance \\
		\hline
		GWP   & $18.11\pm11.82$ & $0.184\pm0.114$  \\
		linear interpolation & $73.12\pm8.26$      &  $11.14\pm2.65$  \\
		log-euclidean & $61.09\pm6.15$       & $9.74\pm1.67$  \\
		\hline
	\end{tabular}%
	\label{tabR2}%
\end{table}%

\subsection{Real DTI field estimated from dMRI}
\begin{figure}[H]
	\centering
	\subfigure[\label{real_t}]{\includegraphics[scale=0.120]{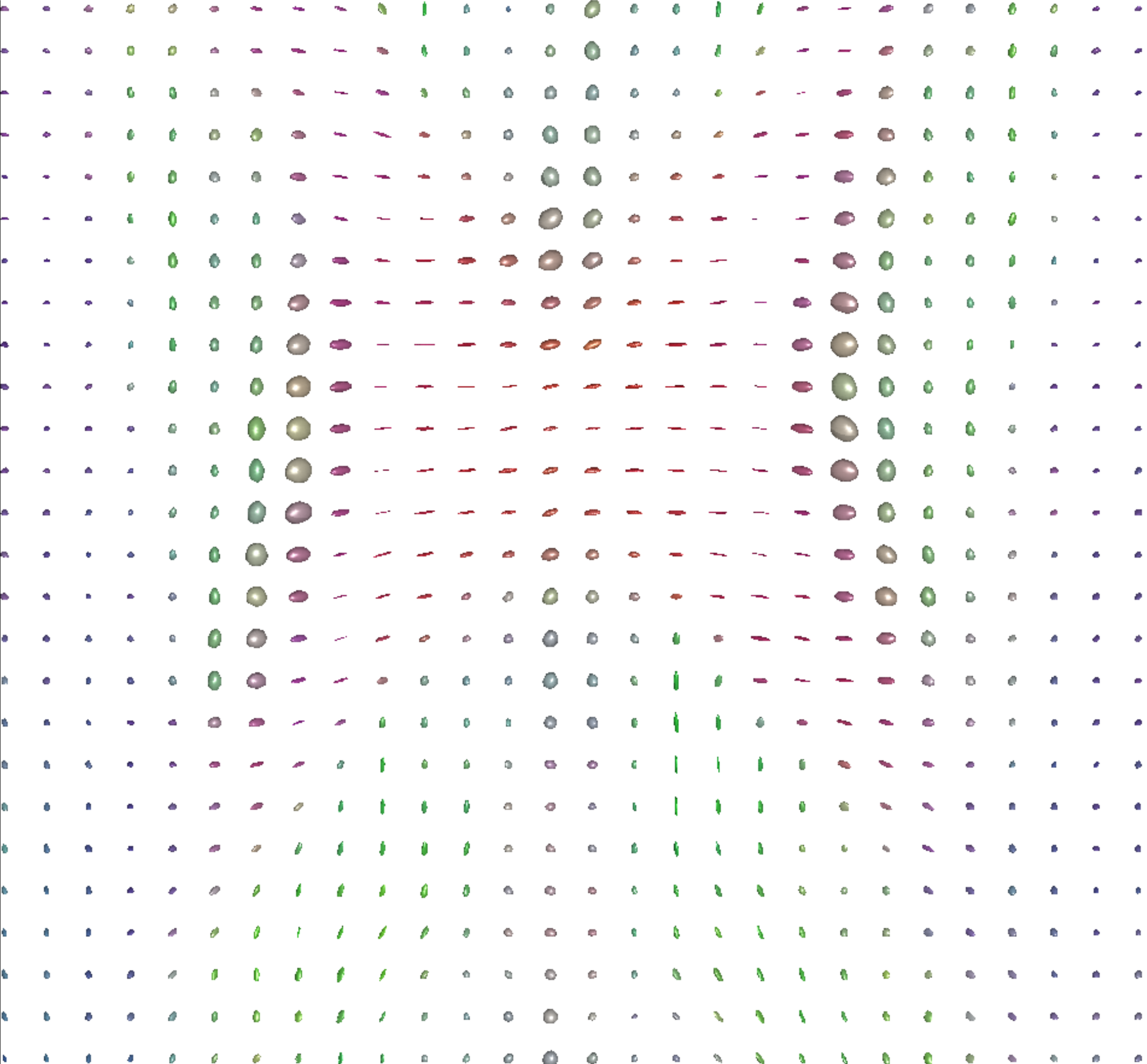}}
	\subfigure[\label{real_lin}]{\includegraphics[scale=0.200]{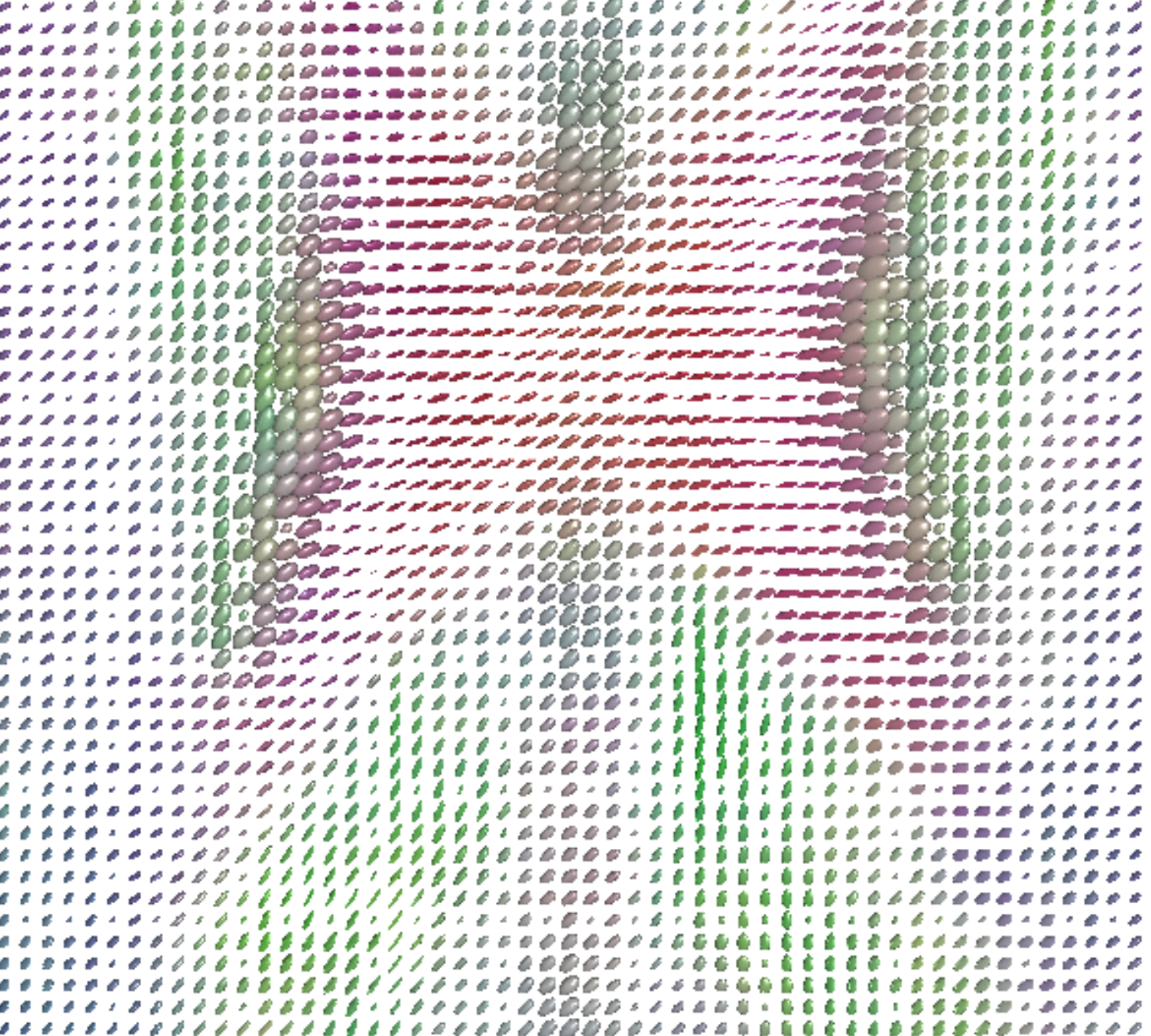}}
	\subfigure[\label{real_log}]{\includegraphics[scale=0.200]{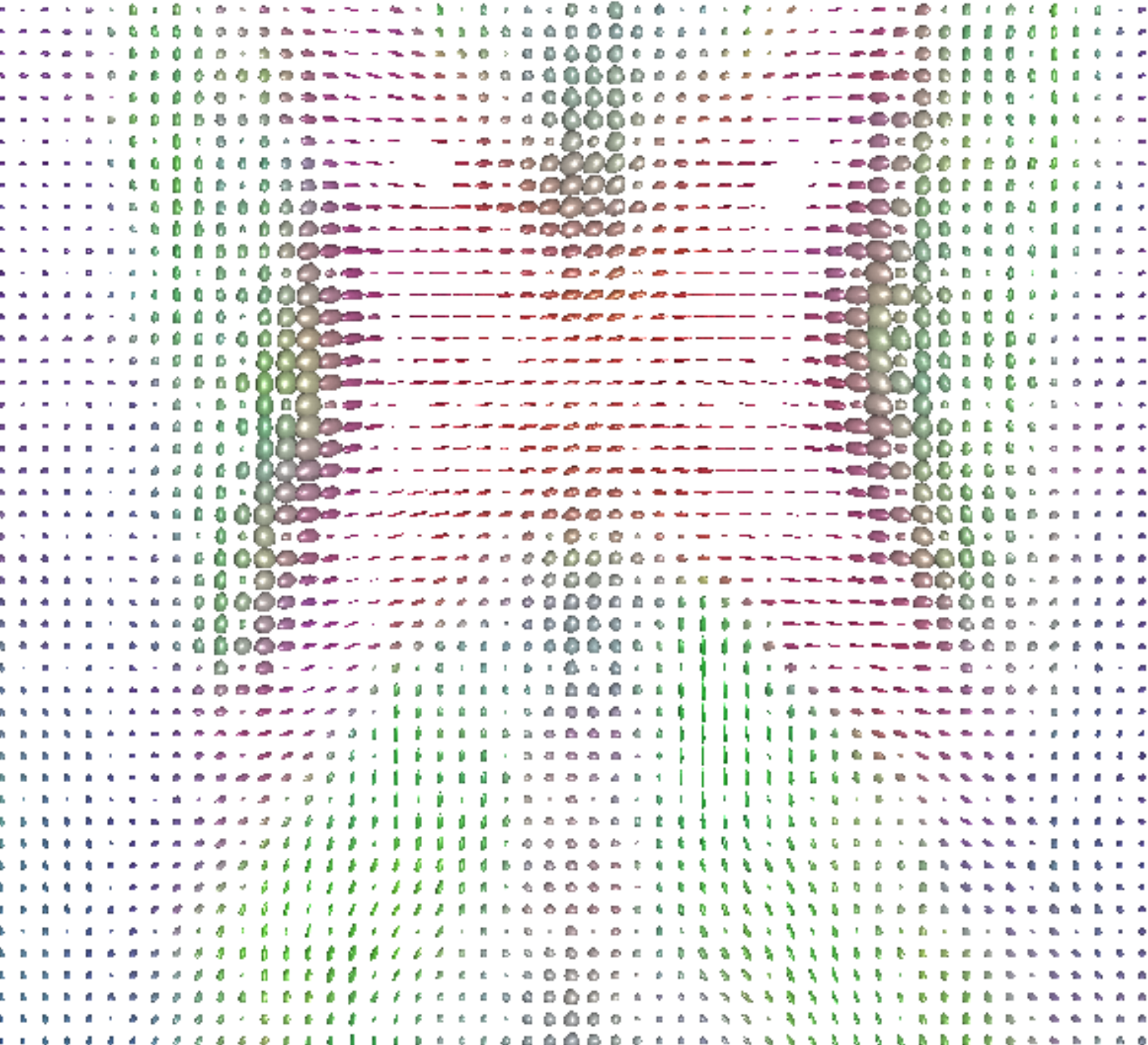}}
	\subfigure[\label{real_gwp}]{\includegraphics[scale=0.120]{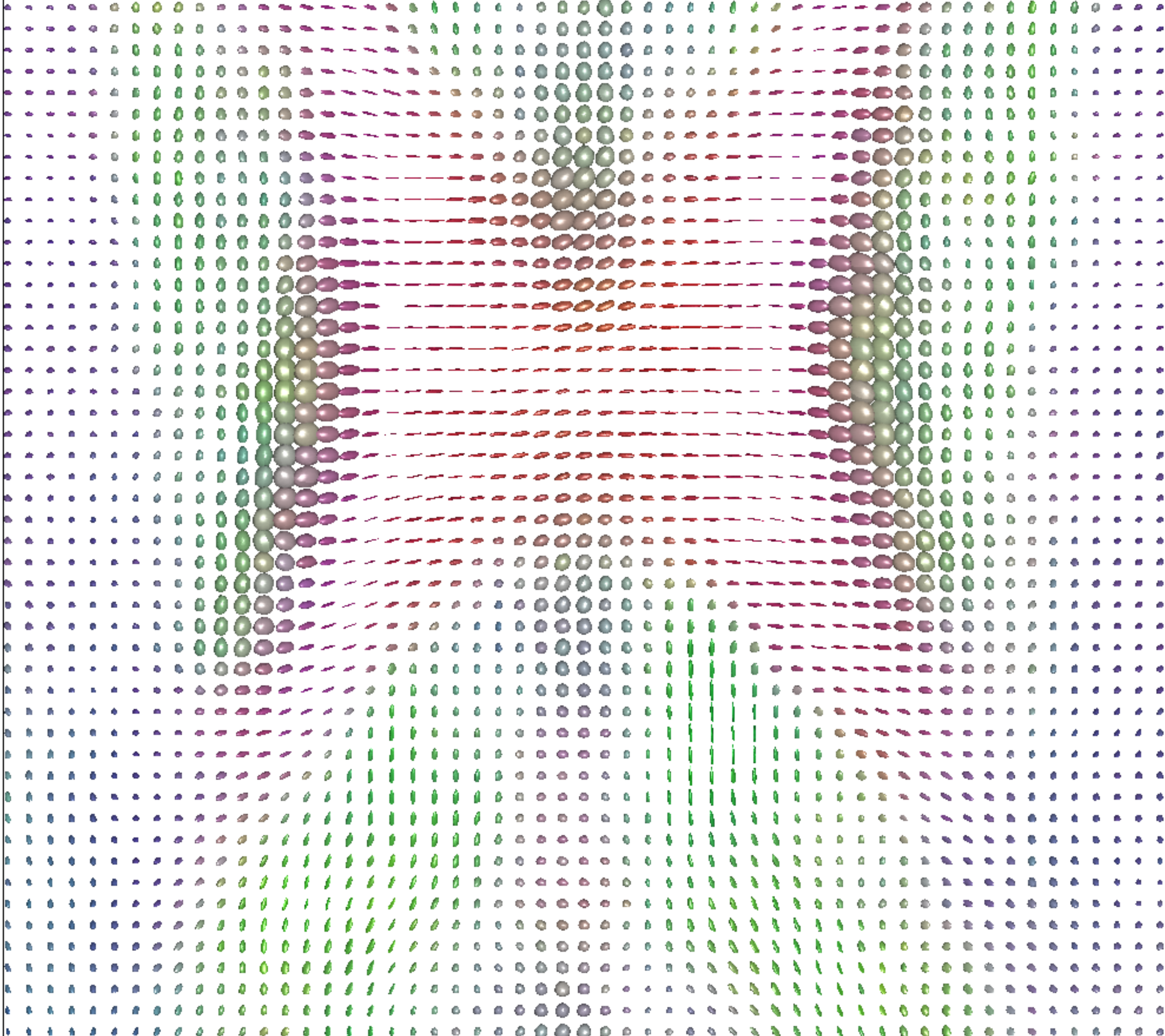}}
	\subfigure[\label{real_gt}]{\includegraphics[scale=0.120]{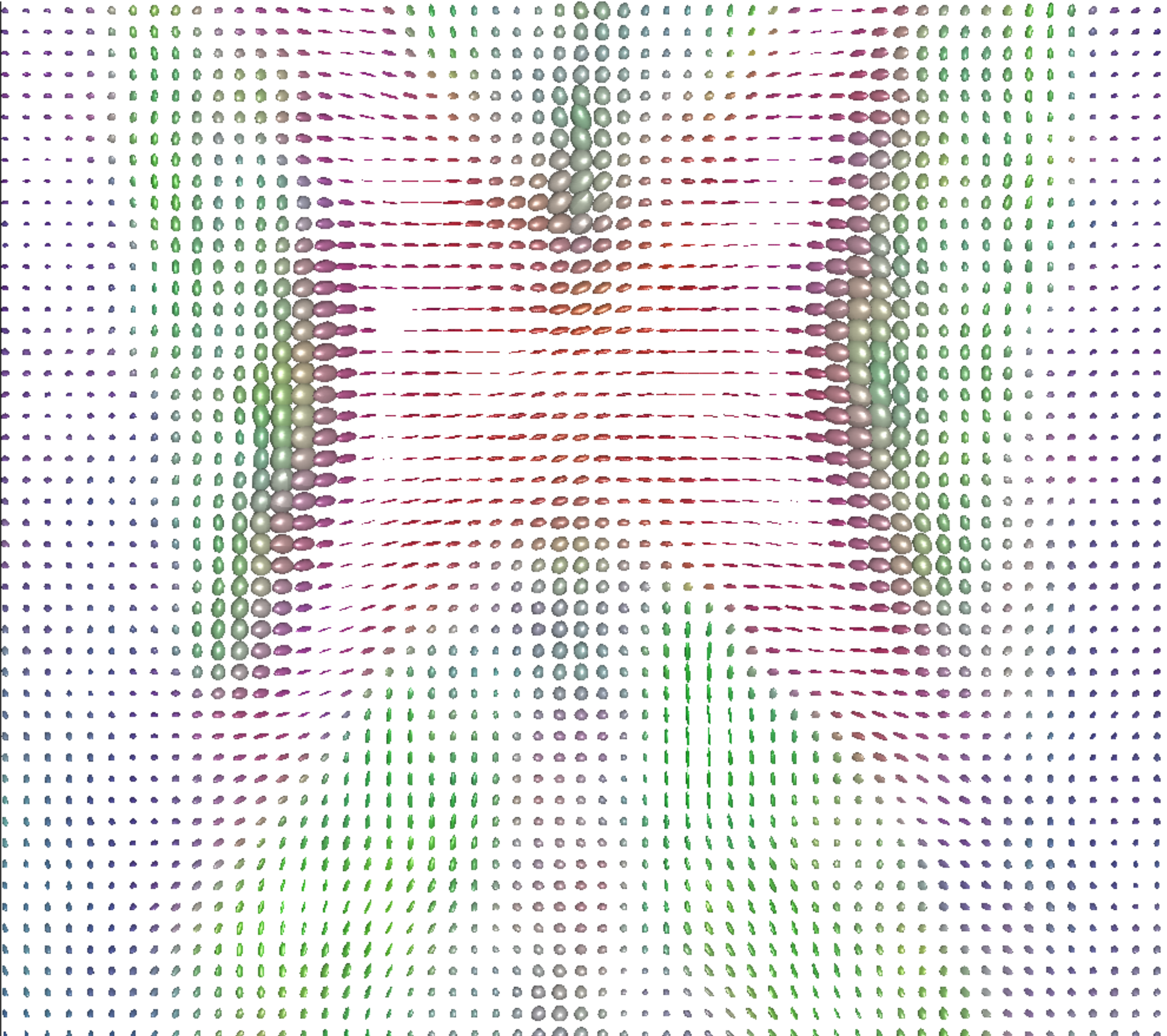}}
	\caption{Graphic results for DTI interpolation $(2\times)$ applied in real DTI data. (a) Downsampled DTI field (data used for estimation). (b) Linear interpolation. (c) Log-euclidean interpolation. (d) Interpolation with GWP. (e) Ground truth.}
	\label{f_real}	
\end{figure}

Finally, we test our method in real DTI data estimated from dMRI
acquired in a human subject. The field corresponds to an axial slice
with $49\times55$ tensors. Figure \ref{f_real} and Table \ref{tabR3}
show the comparative results.

\begin{table}[H]
	\centering
	\caption{Error measures for the real DTI field example}
	\begin{tabular}{ccc}
		\hline
		& Frobenius distance $(\times10^{-5})$ & Riemman distance \\
		\hline
		GWP   & $6.26\pm3.20$ & $0.146\pm0.080$  \\
		linear interpolation &  $45.76\pm7.21$      & $7.25\pm2.10$  \\
		log-euclidean &   $31.67\pm6.10$      & $6.89\pm1.86$ \\
		\hline
	\end{tabular}%
	\label{tabR3}%
\end{table}%

\subsection{Discussion}
Linear and log-euclidean interpolation seek to minimize geodesic distances. The geometric
(Riemann and Euclidean) approaches work well in smooth DTI
fields. However, they reduce their performance in presence of high
level of noise. For example, when we interpolate the synthetic noisy
data (Figure \ref{f_st} and Table \ref{tabR1}), we can observe a
swelling effect for the estimated tensors in the new input locations, when using linear interpolation. 
This is a critical issue, because this effect modifies the fractional anisotropy
maps.  Another big problem with linear interpolation is the
possibility of obtaining non-positive definite tensors. Although,
non-positive definite tensors are avoided with log-euclidean
interpolation, the accuracy of tensor estimation in new input locations is not satisfactory. On the
other hand, GWP guarantees positive definite tensors because of its
mathematical construction. Also, this probabilistic model is more
robust to noise, and it can keep the smooth transition among spatially
nearby data. This property avoids the swelling effect. If we look
at the results for more complex data like crossing fibers DTI, and the real DTI field
(Figures \ref{f_cb},\ref{f_real} and Tables \ref{tabR2},\ref{tabR3}),
we can see better accuracy results for GWP. Both average distances
(Frobenius and Riemann) are smaller in GWP than Linear and
Log-Euclidean methods. Similarly, graphic results of DTI fields are
smoother for GWP interpolation. The GWP takes 
into account the global spatial behavior of the DTI data, while geometric approaches estimate
tensors only with the nearest tensors.

A drawback in GWP happens when there are strong changes in tensor
orientation (i.e. crossing fibers). The GWP cannot capture extreme
modifications in data of crossing fibers. Recall that a GWP is a
superposition of Gaussian processes (GP) and GPs are modeled with
smooth kernels functions. The best alternative to model crossing
fibers is through Higher Order Tensors (HOT). Nevertheless a GWP 
does not describe HOT. 
However, for the geometric approaches, strong
changes in tensor orientation generate worse results than GWP.

In summary, the probabilistic modeling of DTI fields that we employ here
allows a better description of global spatial transition in tensorial
imaging. Geometric approaches are fine for simple DTI fields. However
in real applications, DTI data are very complex (High level of noise,
Heterogeneous data, Non-positive definite tensors, etc). The GWP has
many advantages, for example: it guarantees positive definite tensors,
robustness to noisy data, smooth transition among nearby data, no
swelling effect, it keeps important properties of DTI (FA maps) and
excellent accuracy.

\section{Conclusions and future work}\label{con}
In this paper we developed a probabilistic methodology to interpolate
Diffusion Tensor Imaging (DTI) data. We model a DTI field as a
Generalized Wishart process (GWP). We employ approximate Bayesian
inference for optimizing the relevant variables in GWP. Results obtained
with GWP in synthetic and real DTI data outperform to commonly used
geometric methods. Also, our proposed method guarantees positive
definite tensors, excellent accuracy and it avoids an issue in
tensorial interpolation known as swelling effect. 
As future work, we would like to extend this concept to modeling Higher Order
Tensors. The above approach would be relevant when describing tensorial data from crossing fibers.

\section*{Acknowledgments}

H.D. Vargas Cardona is funded by Colciencias under the program: \textit{formaci\'on de alto nivel para la ciencia, la tecnolog\'ia y la innovaci\'on - Convocatoria 617 de 2013}. This research has been developed under the project 
financed by Colciencias with code 1110-657-40687.

\bibliographystyle{plain}
\bibliography{biblio}

\begin{thebibliography}{10}

\bibitem{Arsigny2006}
V.~Arsigny, P.~Fillard, X.~Pennec, and N.~Ayache.
\newblock Log-euclidean metrics for fast and simple calculus on diffusion
  tensors.
\newblock {\em Mag. Res. Med}, (2):411--421, 2006.

\bibitem{Barmpoutis2010}
A.~Barmpoutis and B.C Vemuri.
\newblock A unified framework for estimating diffusion tensors of any order
  with symmetric positive-definite constraints.
\newblock In {\em Proceedings of ISBI10: IEEE International Symposium on
  Biomedical Imaging}, pages 1385--1388, 2010.

\bibitem{Barmpoutis2007}
A.~Barmpoutis, B.C. Vemuri, T.M. Shepherd, and J.R. Forder.
\newblock Tensor splines for interpolation and approximation of dt-mri with
  applications to segmentation of isolated rat hippocampi.
\newblock {\em IEEE TRANSACTIONS ON MEDICAL IMAGING}, (11):1537--1426, 2007.

\bibitem{Basser1995}
P.J Basser.
\newblock Inferring microstructural features and the physiological state of
  tissues from diffusion weighted images.
\newblock {\em NMR Biomed}, 8:333--344, 1995.

\bibitem{Basser1994}
P.J. Basser, J.~Mattiello, and D.~Le~Bihan.
\newblock Estimation of the effective self-diffusion tensor from the nmr spin
  echo.
\newblock {\em J. Magn. Reson}, 103:247--254, 1994.

\bibitem{Bi2010}
C.~Bi, S.~Takahashi, and I.~Fujishiro.
\newblock Interpolating 3d diffusion tensors in 2d planar domain by locating
  degenerate lines.
\newblock In M.~Abadi and T.~Ito, editors, {\em ISVC'10 Proceedings of the 6th
  international conference on Advances in visual computing}, volume~1, pages
  328--337. Springer-Verlag, 2010.

\bibitem{Chang2012}
I.-S. Chang and X.~Shun-ren.
\newblock Diffusion tensor interpolation profile control using non-uniform
  motion on a riemannian geodesic.
\newblock {\em Comput and Electron}, (2):90--98, 2012.

\bibitem{Fletcher2007}
P.T. Fletcher and S.~Joshi.
\newblock Riemannian geometry for the statistical analysis of diffusion tensor
  data.
\newblock {\em Signal Processing}, (2):250--262, 2007.

\bibitem{Hotz2010}
I.~Hotz, J.~Sreevalsan-Nair, and B.~Hamann.
\newblock Tensor field reconstruction based on eigenvector and eigenvalue
  interpolation.
\newblock {\em Scientific Visualization: Advanced Concepts}, (1):110--123,
  2010.

\bibitem{Kindlmann2007}
G.~Kindlmann, R.S. Estepar, M.~Niethammer, S.~Haker, and C.F. Westin.
\newblock Geodesic-loxodromes for diffusion tensor interpolation and difference
  measurement.
\newblock {\em Med Image Comput Comput Assist Interv}, (1):1--9, 2007.

\bibitem{Restore2005}
C.~Lin-Chin, D.~Jones, and C.~Pierpaoli.
\newblock {R}{E}{S}{T}{O}{R}{E}: Robust estimation of tensors by outlier
  rejection.
\newblock {\em Magnetic Resonance in Medicine}, 53:1088--1095, 2005.

\bibitem{Murray2010}
I.~Murray, R.P. Adams, and D.J. Mackay.
\newblock Elliptical slice sampling.
\newblock {\em JMLR}, 9:541--548, 2010.

\bibitem{Pajevic2002}
S.~Pajevic.
\newblock A continuous tensor field approximation of discrete dt-mri data for
  extracting microstructural and architectural features of tissue.
\newblock {\em J. Magn. Reson.}, pages 85--100, 2002.

\bibitem{ST1965}
J.E Tanner and E.O Stejskal.
\newblock Spin diffusion measurements: Spin echoes in the presence of a
  time-dependent field gradient.
\newblock {\em Journal of Chemical Physiology}, 42:288--292, 1965.

\bibitem{Wilson2011}
A.~Wilson and Z.~Ghahramani.
\newblock Generalised wishart processes.
\newblock In {\em UAI-11}, pages 736--744, 2011.

\bibitem{Yang2014}
F.~Yang, Y.-M. Zhu, J.-H. Luo, J.~Robini, M.~Liu, and P.~Croisille.
\newblock A comparative study of different level interpolations for improving
  spatial resolution in diffusion tensor imaging.
\newblock {\em IEEE Journal of Biomedical and Health Informatics},
  (4):1317--1327, 2014.

\bibitem{Yang2012}
F.~Yang, Y.-M. Zhu, I.E. Magnin, J.-H. Luo, P.~Croisille, and P.B. Kingsley.
\newblock Feature-based interpolation of diffusion tensor fields and
  application to human cardiac dt-mri.
\newblock {\em Medical Image Analysis}, (1):459--481, 2012.

\end{thebibliography}

\end{document}